\documentclass[a4paper, 10pt, conference]{ieeeconf}      
\usepackage{FG2020}
\usepackage{times}
\usepackage{epsfig}
\usepackage{graphicx}
\usepackage{amsmath}
\usepackage{amssymb}

\FGfinalcopy 

\IEEEoverridecommandlockouts                              
\def\FGPaperID{45} 

\title{\LARGE \bf
DeeSCo: Deep heterogeneous ensemble with Stochastic Combinatory loss for gaze estimation
}

\author{\parbox{16cm}{\centering
    {\large Edouard Yvinec$^1$ and Arnaud Dapogny$^1$ and K\'evin Bailly$^{1,2}$}\\
    {\normalsize
    $^1$ Datakalab, 114 Boulevard Malesherbes, 75017 Paris, France\\
    $^2$ ISIR, Sorbonne Universit\'e, 4 place Jussieu, 75005 Paris, France}}
}

\begin{document}

\ifFGfinal
\thispagestyle{empty}
\pagestyle{empty}
\else
\author{Anonymous FG2020 submission\\ Paper ID \FGPaperID \\}
\pagestyle{plain}
\fi
\maketitle

\begin{abstract}

From medical research to gaming applications, gaze estimation is becoming a valuable tool. While there exists a number of hardware-based solutions, recent deep learning-based approaches, coupled with the availability of large-scale databases, have allowed to provide a precise gaze estimate using only consumer sensors. However, there remains a number of questions, regarding the problem formulation, architectural choices and learning paradigms for designing gaze estimation systems in order to bridge the gap between geometry-based systems involving specific hardware and approaches using consumer sensors only. In this paper, we introduce a deep, end-to-end trainable ensemble of heatmap-based weak predictors for 2D/3D gaze estimation. We show that, through heterogeneous architectural design of these weak predictors, we can improve the decorrelation between the latter predictors to design more robust deep ensemble models. Furthermore, we propose a stochastic combinatory loss that consists in randomly sampling combinations of weak predictors at train time. This allows to train better individual weak predictors, with lower correlation between them. This, in turns, allows to significantly enhance the performance of the deep ensemble. We show that our Deep heterogeneous ensemble with Stochastic Combinatory loss (DeeSCo) outperforms state-of-the-art approaches for 2D/3D gaze estimation on multiple datasets.

\end{abstract}

\section{INTRODUCTION}

Gaze estimation is an important computer vision research field, as it serves a broad spectrum of applications such as entertainment on massive online streaming platforms \cite{kruikemeier2018learning}, social networks and can also be used in order to refine psychological analysis by correlating behavior and sensory stimulation \cite{ishii2016prediction, huang2016stressclick, yarbus1967eye}, as well as surgery \cite{james2007eye}. A review on gaze estimation can be found in \cite{kar2017review}. 

\begin{figure}[ht]
\centering
\includegraphics[width = \linewidth]{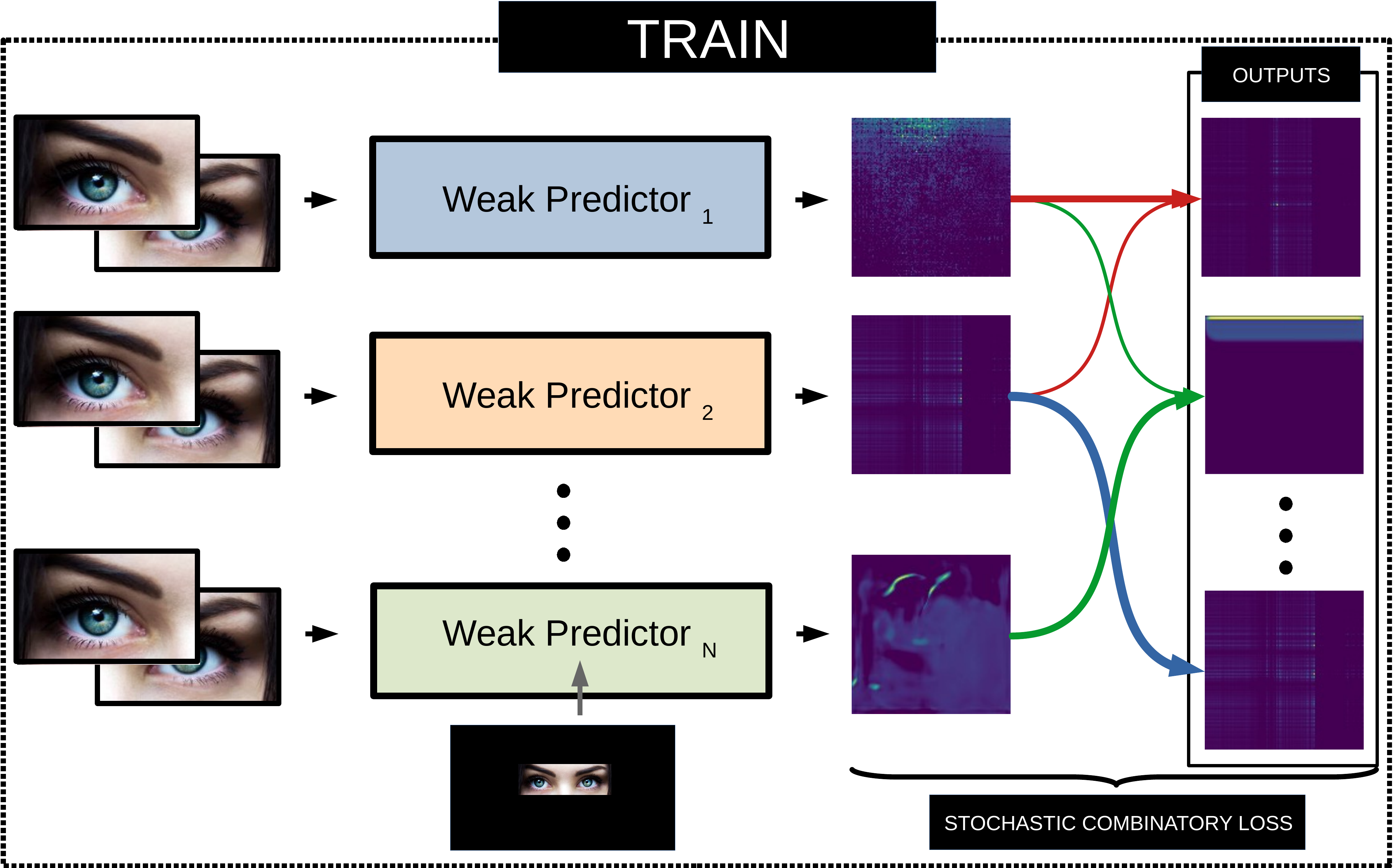}
\caption{\textbf{Overview of the proposed system:} at train time (left), several weak predictors with heterogeneous architectural designs are trained with a stochastic combinatory loss to learn decorrelated individual weak predictors. The deep heterogeneous ensemble of these $N$ weak predictors (right) thus outputs an accurate gaze 2D/3D gaze estimation.}
\label{mainfig}
\end{figure}

Generally speaking, gaze estimation methods can be broken down into 2 different kinds of approaches. On the one hand, the so-called geometric approaches based on specific hardware rely on a spherical modelization of the eye to infer the 3D localization of the eyeball center as well as the iris from a number of 2D landmarks in the eye region (e.g. the inner and outer corner of the eye and the center of the iris \cite{ishikawa2004passive}). These methods usually rely on the evaluation of anatomical constants, among which the radius of the eye, the distance between the center of the eye and the orthogonal plane of the eye corners. In other instances, the eyeball is represented as two nested spheres \cite{ohno2002freegaze, park2018learning}. As such they offer the advantage of not being dependent on large databases for learning the prediction models. However, such approach usually requires that these landmarks can be localized very precisely on the image, especially in the case where the underlying eye model is very complex. Thus, these approaches requiring dedicated hardware such as very high-resolution cameras or infra-red sensors to register the eye region. Nevertheless, given such specific device and in nominal conditions, these methods can achieve a 2D gaze estimation error of about $\approx 1$ cm, which is very low. On the other hand, appearance-based methods only consider the raw face image, or crops extracted around the location of the eyes. Alternatively, additional information such as head-pose, facial landmarks and relative crop position on the original picture can also be used. This raw information is then passed through a deep neural network that aims at directly mapping it to a 2D or 3D gaze estimation. These methods however have two drawbacks: First, as most deep learning-based approaches, they rely upon having large databases. Second, the accuracy of these methods is currently limited, as the best performing model for on-screen 2D gaze estimation \cite{sugano2017s} has an accuracy of $\approx 4$ cm. Thus, appearance-based methods are still significantly less accurate than hardware-based methods.

While the gap between hardware-based and appearance-based method remains non-negligible, recent breakthrough in the understanding of deep neural networks, combined with the availability of large-scale gaze estimation databases bring hope for designing accurate gaze prediction systems using solely consumer sensors. Towards this aim, ensembling deep models constitute an ideal solution for increasing the accuracy of deep learning-based models while reducing their variance. However, how to effectively design deep ensembles is a largely overlooked problem in the literature. In this paper, we present DeeSCo, a \textbf{Dee}p ensemble method for gaze estimation, trained with \textbf{S}tochastic \textbf{Co}mbinatory loss, illustrated on Figure \ref{mainfig}. We show that through heterogeneity in the weak predictors architectures, we can induce decorrelation between predictions of these predictors, enhancing the ensemble accuracy. Furthermore, we propose a novel stochastic combinatory loss to enhance the robustness of the weak predictors, as well as to increase the decorrelation between them, enhancing the ensemble accuracy. The key contributions of this paper are thus three-fold:

\begin{itemize}
\item We introduce a deep ensemble of models for heatmap-based 2D/3D gaze estimation. We show that through heterogeneous architectural design we increase the decorrelation between the weak predictors, which is key to the performance of the deep ensemble.
\item We propose a stochastic combinatory loss to ensure that each randomly-sampled weighted combination of weak predictors outputs relevant prediction. This increases both the individual weak predictor accuracy as well as the decorrelation between these, which in turn significantly enhances the ensemble prediction.
\item Through experimental validation on several databases on both 2D and 3D gaze estimation, we show that DeeSCo significantly outperforms the state-of-the-art.
\end{itemize}

\section{Related work}
Geometric based approaches, relying on dedicated hardware and calibration, use eye features such as pupil center, iris contours or even stereo and depth cameras as described in \cite{jianfeng2014eye,reale2011multi,shih2004novel}. More specifically, \cite{jianfeng2014eye} uses RGB-D Kinect sensor in order to predict the gaze by setting up a model with the help of a calibration phase where an user shall look at an unpredefined target. \textit{Reale et al.} \cite{reale2011multi} uses a two-camera system to track head, eyes, and mouth movements. Similarly, \cite{shih2004novel} employs multiple cameras and multiple light sources to estimate the optical axis of a user’s eye without any user-dependent parameters or calibration phase.

Other geometrical methods use shape parameters extracted from high resolution, eye images such as boundary of eye limbus and iris in order to associate it with a fixed human eye model, as in \cite{baltruvsaitis2016openface, valenti2011combining, wood2014eyetab}. \textit{Baltru{\v{s}}aitis et al.} \cite{baltruvsaitis2016openface} employ a deformable part model shape registration approach to detect eye-region landmarks and then estimate gaze direction from said landmarks. Authors in \cite{valenti2011combining} propose a hybrid scheme combining head pose and eye location information to predict gaze while \textit{Wood et al.} \cite{wood2014eyetab} get eye positions from the image using a cascade of classifiers in order to determine the limbus’ elliptical outline before estimating gaze for each eye.

Due to recent development of deep learning in computer vision, the interest in appearance-based methods has significantly grown, essentially due to the use of CNN's \cite{lu2011inferring,sugano2014learning, zhang2015appearance, cvpr2016_gazecapture}.
For instance, authors in \cite{sugano2014learning, yu2019improving} make use of synthetic data so as to refine predictions. \textit{Wang et al.} \cite{wang2019generalizing} propose the use of adversarial learning to address cross-database and cross-subject generalization discrepancies in the models, as well as a bayesian hyperparameter optimization scheme. The work in \cite{xiong2019mixed} propose to better use the underlying structure of existing gaze estimation databases, where a large number of non-i.i.d. samples coming from few subjects can be found. Many such approaches employ generative adversarial networks \cite{wang2019generalizing, sela2017gazegan,shrivastava2017learning}. Last but not least, as it was done in some geometric-based methods, a number of approaches investigate the use of person-specific calibration \cite{lu2011inferring,liu2019differential,park2018learning}. Such approach usually yields good results, at the price of a lesser convenience.

Another way to increase the accuracy of deep models without involving such calibration phase is to use ensemble methods. Ensemble methods \cite{dietterich2000ensemble} are ubiquitous and widely studied models in the machine learning community. However, much less studied is the use of deep ensembles in general, and for gaze estimation in particular. A good general rule of thumb in ensemble methods \cite{breiman2001random} is that the accuracy of the ensemble is proportional to the individual predictive capacity of the weak predictors that constitute it, as well as the diversity between the latter, or the decorrelation that exists between the predictions outputted by these weak predictors. In \cite{furlanello2018born}, authors recommend to use ensembles of generations of teacher-student models to improve the overall predictive accuracy. Authors in \cite{song2018collaborative} propose a collaborative learning scheme to put emphasis on the complementary aspect of the weak predictors. As such, authors in \cite{lee2016stochastic} introduce a stochastic multiple choice learning scheme to promote diversity of the ensemble. Contrarily to these works, we study how bringing architectural diversity among the weak predictors can enhance the accuracy. Also, as deep ensembles are generally made of small numbers of deep models, we can explicit all the possible combinations in what we call our stochastic combinatory loss.

The rest of the paper is organized as follows: in Section \ref{methodo}, we present our DeeSCo model, including deep heterogeneous ensemble and stochastic combinatory loss. In Section \ref{expes} we validate our approach for 2D/3D gaze estimation and provide insight on the model behavior. Finally, in Section \ref{concl} we conclude and discuss future perspectives.

\section{Methodology overview}\label{methodo}

In this section, we present our DeeSCo method for gaze estimation. We first show that 2D/3D gaze can be estimated from a heatmap. In Section \ref{individualnets} we propose a number of innovative architectural solutions to generate such heatmap using deep neural networks. In Section \ref{ensemblenets} we ensemble such heterogeneous models and in Section \ref{lossoverview} we introduce our combinatory loss to better implement complementarity between the individual weak models. First, 2D gaze estimation refers to predicting where a person looks on a target screen, thus the gaze position can trivially be converted into a normalized frame, e.g. $[-1;1]^2$. As for 3D gaze, the ground truth gaze can be represented by its yaw ($\gamma$) and pitch ($\beta$):

$$ u^{*{}} = \left(\begin{array}{c}
cos(\gamma^{*{}}) sin(\beta^{*{}}) \\
sin(\gamma^{*{}}) \\
cos(\beta^{*{}})^2
\end{array} \right) $$

Thus, in this case, we can also use a heatmap-generating model to provide an estimate $\hat{v}=(\hat{\gamma},\hat{\beta})$ that shall match the yaw-pitch ground truth vector $v^{*{}}=(\gamma^{*{}},\beta^{*{}})$. This kind of conversion is convenient because, as pointed out in \cite{liu2018intriguing}, translational invariance created by using a number of CNN layers for feature extraction can hinder the accuracy of deep networks for localization tasks. Because of this, providing prediction under the form of a heatmap provides a simple solution to incorporate the nature of the gaze estimation task as a spatial localization one into a deep framework.

\subsection{Heatmap-based gaze estimation}
\label{individualnets}

\begin{figure}
\centering
\includegraphics[width = 0.95\linewidth]{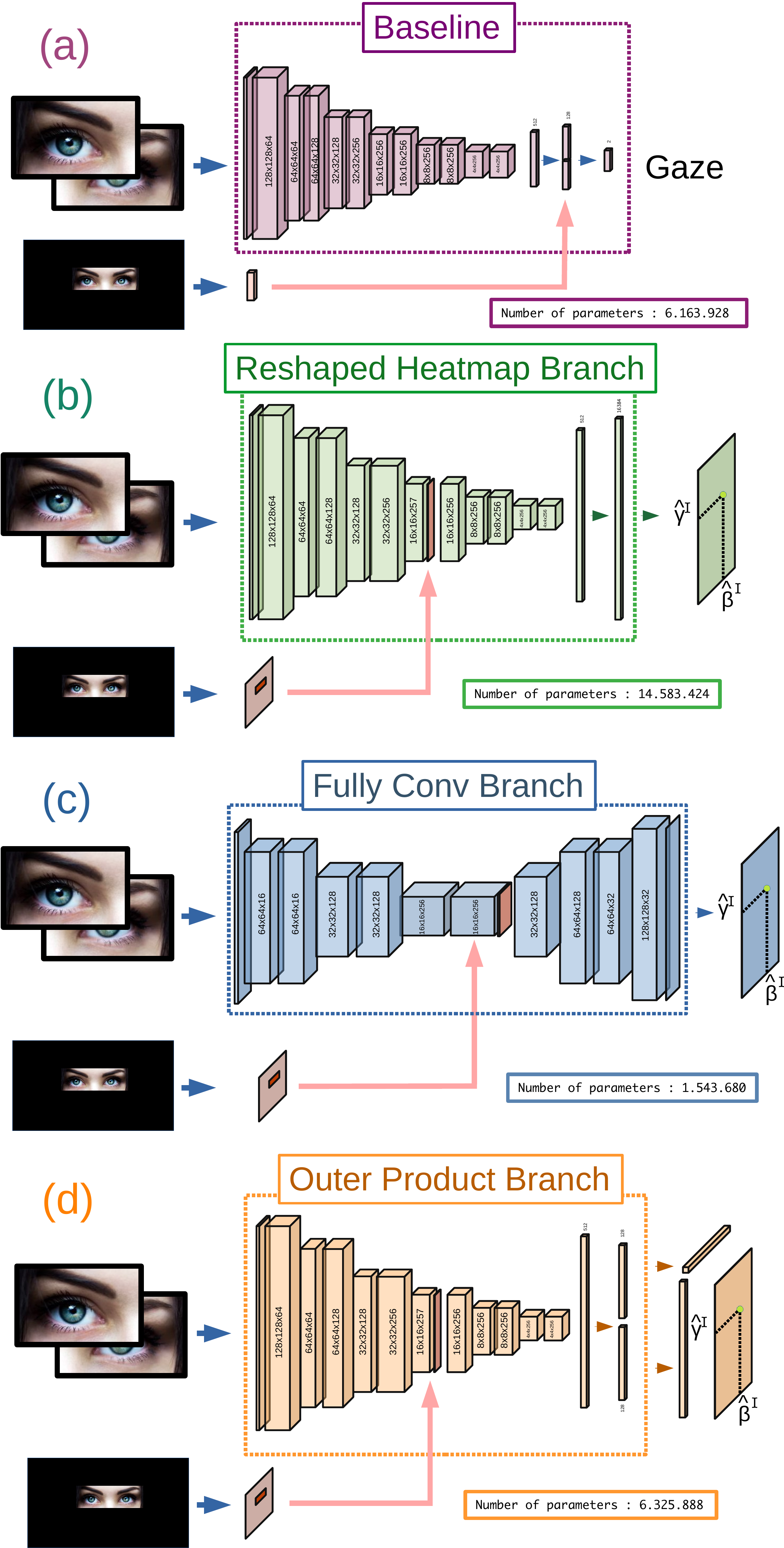}
\caption{Different architectures for heatmap-based gaze weak predictors.}
\label{gallmodels}
\end{figure}

Figure \ref{gallmodels} provides an outline of different architectures that can be used for heatmap-based gaze prediction. For each of these models, crops corresponding to the two eyes are provided to CNN layers. Similarly to Krafka \textit{et al.} \cite{krafka2016eye} the localization of the crop within the face is provided as a $16 \times 16$ binary mask and concatenated to the feature maps at the same size. For each weak model indexed by $i$, a heatmap $H_i$ is generated by applying either of the following architecture:

\begin{itemize}

\item \textbf{The baseline (non-heatmap) branch (Figure \ref{gallmodels}-a)} is obtained by applying CNN and fully-connected (FCN) layers, which directly maps to a 2-dimensional output.

\item \textbf{The reshaped heatmap branch (Figure \ref{gallmodels}-b)} is obtained by applying a large FCN layer, which is then reshaped into a $128 \times 128$ heatmap. Because of these FCN, this weak model embraces lots of parameters.

\item \textbf{The fully-convolutional branch (Figure \ref{gallmodels}-c)} is obtained by applying transposed convolutions to upsample the $16 \times 16$ feature maps and using a single $1 \times 1$ CNN to convert the final embedding to a final, full-resolution, $128 \times 128$ heatmap. Also note that because the structure of the input eye crops and output yaw-pitch gaze heatmaps are largely unrelated in their spatial structure, we do not use skip-connection as in U-net architecture. This weak predictor also have a lower number of parameters due to the absence of FCN layers.

\item \textbf{The outer product branch (Figure \ref{gallmodels}-d)} is similar to the reshaped heatmap branch in its architecture as it also involves transformation using FCN layers, however this branch outputs two separate outputs $\textbf{a}$ (marginal vertical vector) and $\textbf{b}$ (marginal horizontal vector), upon which a heatmap is generated by applying an outer product operator $H_i=\textbf{a} \otimes \textbf{b}$. Compared to the reshaped heatmap branch, this model has a dramatically lower number of parameters, as it significantly reduces the number of neurons in the FCN layers.
\end{itemize}

\subsection{Model Ensembling}\label{ensemblenets}

We merge together the outputs of $N$ weak models with potentially different architectures, each of these weak models taking as input both the eye crops as well as the crop position, as highlighted previously. We use a $1 \times 1$ convolution to learn an adaptive mixture $H$, \textit{i.e.} $\forall (x,y) \in [0,127]^2$:

\begin{equation}
H(x,y)=\sum_{i=1}^N \lambda_i H_i(x,y)
\end{equation}

This merged heatmap can then be converted to a probability map by applying spatial softmax:

\begin{equation}
\phi(x,y)=\frac{\exp(H(x,y))}{\sum_x \sum_y \exp(H(x,y))}
\end{equation}

The position of the maximum of $\phi(x,y)$ in yaw-pitch space can be obtained by computing the first-order moments of $\phi(x,y)$ w.r.t. the horizontal and vertical axis:

\begin{equation}\label{fullmodel}
\left \{
\begin{array}{l @{=} l}
\hat{\gamma}(H)&\mathbb{E}_{x,y}[x\phi(x,y)] \\
\hat{\beta}(H)&\mathbb{E}_{x,y}[y\phi(x,y)] \\
\end{array}
\right.
\end{equation}

This trick for providing a differentiable estimate of the argmax operator can be found in \cite{luvizon2017human}. By using such model, the deep ensemble (\textit{i.e.} the weak predictors and the heatmap merging CNN layer) can be trained jointly, in an end-to-end manner. As pointed out in the seminal work of Breiman \cite{breiman2001random}, the success of ensemble models stems from (a) the predictive accuracy of individual weak predictors, and (b) the decorrelation between these predictors. In practice, a classical way to decorrelate these models (similarly to random subspace methods \cite{breiman2001random}) would be to provide different inputs or fractions thereof, e.g. training weak models with only one eye crop. In what follows, we show that contrary to what is commonly done in the gaze tracking literature, this does significantly lower the prediction accuracy. Nevertheless, we propose another way to promote diversity among the weak predictors, which consists in using different architectures, e.g. the ones introduced in Section \ref{individualnets}. Furthermore, we introduce a combinatory loss that uses various combinations of weak models in training to more efficiently learn complementarity between these predictors.

\subsection{Stochastic Combinatory loss}\label{lossoverview}

\begin{figure}
\centering
\includegraphics[width = \linewidth]{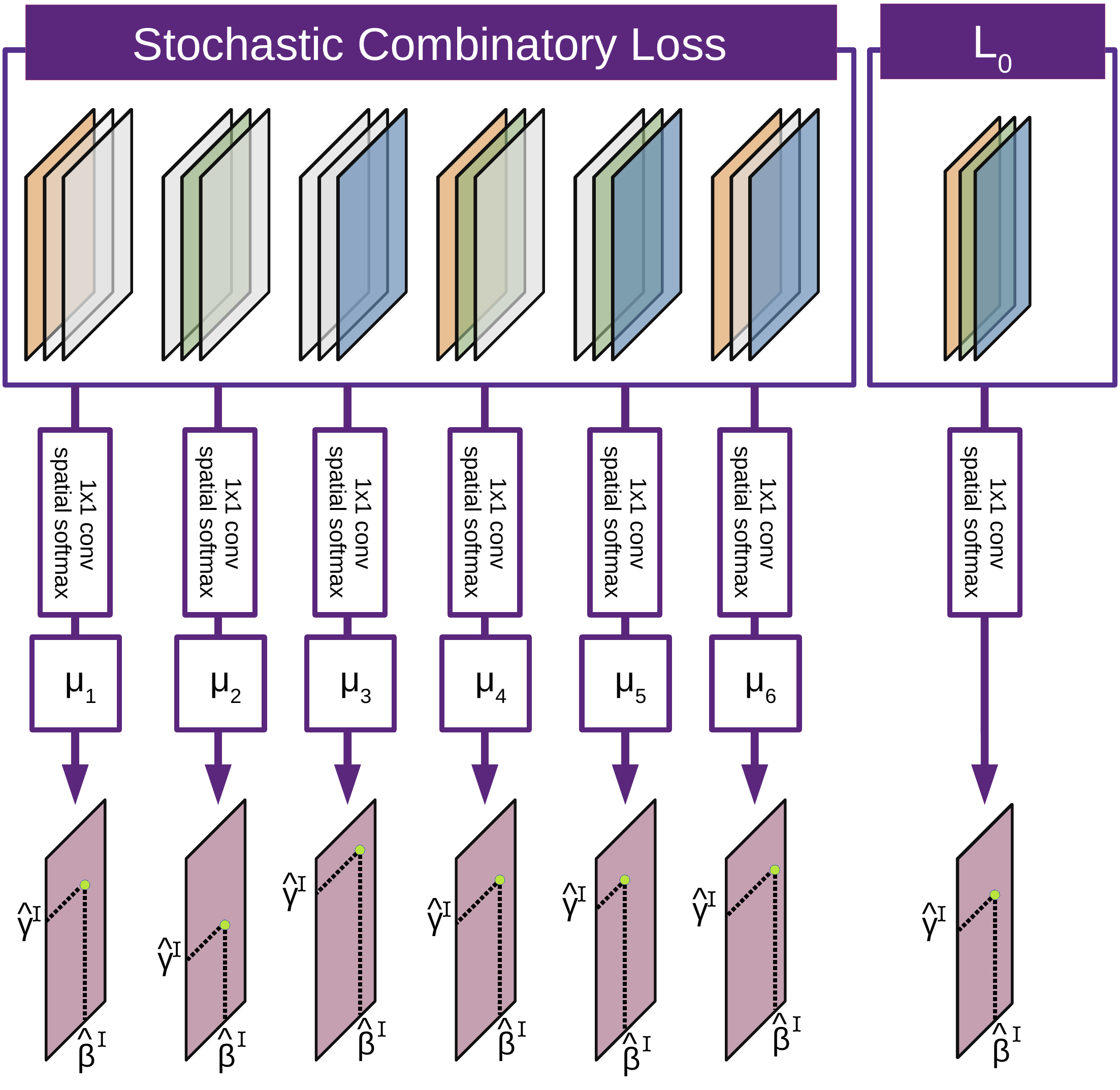}
\caption{Illustration of our combinatory loss with $N=3$ weak models. For each combination $I$ of weak models we draw a random weight $\mu^I$ for each batch, increasing the decorrelation and the accuracy of each combination.}
\label{lossfigure}
\end{figure}

As suggested in previous section, when optimizing a loss function $\mathcal{L}_0 =(\hat \gamma-\gamma^{*{}})^2+(\hat \beta-\beta^{*{}})^2$ (e.g., in this case, $\mathcal{L}_0$ is a $\mathcal{L}_2$ loss) over the ensemble, some weak predictors may end up learning specific prediction instead of gaze estimation. To address this problem, we may enforce that each possible combination of weak predictors shall be, to a certain extent, relevant for gaze prediction. Figure \ref{lossfigure} provides an outline of our loss for $N=3$. Let $\mathcal{I}$ denotes the set of all strict non-empty subparts  of $\mathopen{[}0\,;N-1\mathclose{]}$. We have $\text{card}(\mathcal{I})=\sum_{n=1}^{N-1}\binom{N}{n}=2^N-2$. For each element $I=\{i_1,...,i_n\} \in \mathcal{I}$ with $n$ element we add a $1 \times 1$ CNN layer with weights $\{\lambda_{i_1}^I,...,\lambda_{i_n}^I\}$. The output probability map of the sub-ensemble using only models $\{i_1,...,i_n\}$ can be written as:

\begin{equation}
H^I=\sum_{j=i_1,...,i_n} \lambda_{j}^I H_j(x,y)
\end{equation}

And the yaw and pitch estimates $\hat{\gamma}^I = \hat{\gamma}(H^I)$ and $\hat{\beta}^I = \hat{\beta}(H^I)$ defined as in Equation \eqref{fullmodel}. We then define the loss corresponding to this partial combination of models $\mathcal{L}^I =(\hat \gamma^I-\gamma^{*{}})^2+(\hat \beta^I-\beta^{*{}})^2$. We then consider a random convex combination of weights $\{\mu^I\}_{I \in \mathcal{I} }$ obtained by random sampling from a unit interval followed by normalization. Our combinatory loss $\mathcal{L}_{\text{comb}}$ can be expressed as :

\begin{equation}
\mathcal{L}_{\text{comb}}=\sum_{I \in \mathcal{I}} \mu^I \mathcal{L}^I
\end{equation}

We draw new values $(\mu^I)_{I \in \mathcal{I} }$ for every batch while the values of $\lambda^I$ are updated via backprop. The final loss $\mathcal{L}_{\text{tot}}$ is written as a weighted combination of the standard loss and the proposed combinatory loss:

\begin{equation}
\mathcal{L}_{\text{tot}}=\mathcal{L}_{0}+\nu \mathcal{L}_{\text{comb}}
\end{equation}

With $\nu$ a hyperparameter balancing the contributions of the two losses. This loss allows each \textit{combination} of weak models to output decent gaze predictions. From an ensemble perspective, the random weighting procedure allows to disentangle the contributions of each sub-model, effectively creating decorrelation while enhancing their individual capacity.


\section{Experiments}\label{expes}

In this section, we first introduce the 2D/3D gaze estimation datasets (Section \ref{datasets}) upon which we validate our models. In Section \ref{impldetails} we present implementation details of our method to ensure reproducibility of the results. Then, in Section \ref{3dgaze} we validate our DeeSCo model for 3D gaze estimation through extensive ablation study of the proposed deep heterogeneous ensemble and combinatory loss, as well as through comparisons with current state-of-the-art approaches. In Section \ref{2dgaze} we show that DeeSCo also outperforms other existing methods for 2D gaze estimation. Lastly, in Section \ref{modelintro} we provide qualitative insight on the importance of our stochastic combinatory loss for learning meaningful heatmaps.

\subsection{Datasets}\label{datasets}
We evaluated our models on two datasets, namely the MPIIGaze and UT Multiview datasets.

The \textbf{MPIIGaze}  \cite{zhang2015appearance} dataset offers a great variety of head-poses and illumination over 213659 examples coming from 15 subjects. The dataset consists of cropped images (the full face corresponding image is in the face data-set MPIIFace \cite{sugano2017s}). As it is traditionally done in the literature, we report the angular error (for 3D gaze estimation) and Euclidean error (for 2D gaze estimation) in a leave-one-subject-out cross-validation scheme.

The \textbf{UT Multiview} \cite{sugano2014learning} dataset contains $\approx 64000$ images coming from 50 different subjects, each recording with 8 views and 160 gaze directions. As it is classically done, we report the angular error in a 3-folds subject-independent cross-validation setting on this dataset.

\subsection{Implementation details}\label{impldetails}

We used an ensemble composed of $N=3$ weak predictors as described in Figure \ref{gallmodels}. Our model takes as inputs, a normalized cropped image of each eye of shape $128 \times 256 \times 3$ and $16\times 16$ map representing the relative position of the eyes in the original image.

Each branch is composed of a number of $3 \times 3$ convolutional layers, each followed with batch normalization and ReLU activation. The number of feature maps for each layer is illustrated on Figure \ref{gallmodels}. Optimization is performed by applying ADAM Optimizer with $\beta_1=0.9$. We use a base learning rate of $2.10^{-4}$ with polynomial annealing and trained using batches of 32 elements over 10 epochs per fold for both MPIIGaze and UT Multiview benchmarks, for a total of about $60000$ steps per fold on the leave one out evaluation on MPIIGaze dataset and $20000$ steps per fold on a 3-fold evaluation on UT Multiview dataset.

\subsection{3D gaze estimation}\label{3dgaze}

\subsubsection{Architecture Ablation}

\begin{table}
\begin{center}
\caption{Ablation study for combinations of different architectures. \textbf{Ba:} baseline branch, \textbf{Ou:} Outer product branch, \textbf{Rh:} reshaped heatmap branch, \textbf{Fc:} Fully convolutional branch. All models are trained for 4 epochs for each cross-validation fold.}
\label{tab:3D_models_ablation_study_MPII}
\begin{tabular}{|c|c|c|c|c|}
\hline
 Branch 1         & Branch 2        & Branch 3        & err (deg)             & \#params  \\
\hline\hline
\multicolumn{5}{|c|}{MPII}\\
\hline
\hline
Ba (l+r)         & -        & -        & 5.33                 & 6.1M   \\
Ba (l+r)         & Ba (l+r)        & -        & 5.37           & 12.3M   \\
Ba (l)     & Ba (r)    & -        & 5.34                 & 12.3M   \\
Rh (l)     & Rh (r)    & -        & 5.15                 & 29M   \\
Ou (l)     & Ou (r)    & -        & 5.26                 & 13M   \\
Rh (l+r)  & Rh (l+r) & -        & 4.78                 & 29M   \\
Rh (l+r)  & Rh (l+r) & Rh (l+r) & 4.89                 & 43.5M   \\
Rh (l+r)  & Rh (l+r) & Fc (l+r) & 4.43                 & 30.5M   \\
Rh (l+r)  & Ou  (l+r) & Fc (l+r) & 4.46                 & 22.3M   \\
\hline
\hline
\multicolumn{5}{|c|}{UT Multiview}\\
\hline
\hline
Rh (l+r)  & Rh (l+r) & -        & 5.41                 & 29M   \\
Rh (l+r)  & Rh (l+r) & Rh (l+r) & 7.24                 & 43.5M   \\
Rh (l+r)  & Rh (l+r) & Fc (l+r) & 5.01                 & 30.5M   \\
Rh (l+r)  & Ou  (l+r) & Fc (l+r) & 5.04                 & 22.3M   \\
\hline
\end{tabular}
\end{center}
\end{table}

In order to speed-up training, we learned our models over $4$ epochs to retain the best deep ensemble architecture for gaze estimation. We display the results on MPIIGaze and UT multiview datasets in Table \ref{tab:3D_models_ablation_study_MPII}. First, we show that we can obtain the same accuracy on MPII by using a baseline network (Ba) that takes as input the two eye crops in the same network, as opposed to duplicating a similar architecture that processes the two eyes independently, which is a popular approach for eye tracking \cite{cvpr2016_gazecapture}. This allows to substantially reduce the number of parameters and increase computational efficiency. Also, using reshaped heatmap (Rh) branches instead of the baseline branch (Ba) is more efficient at the cost of adding more hyperparameters. This can, however, be mitigated by using outer product (Ou) branches instead. Furthermore, even though ensembling models tends to increase the overall accuracy, this is not true if we stack similar architectures (e.g. an ensemble of 3 Rh branches is worse that 2). If, however, we design heterogeneous ensembles (Rh+Rh+Fc) we obtain higher accuracies with lower total parameter counts. The trend is similar on UT multiview dataset, where the best accuracy is provided by a Rh-Rh-Fc ensemble. Last but not least, using a fully-heterogeneous ensemble such as Rh-Ou-Fc allows to obtain very similar accuracy with lighter architectures, as the decrease in individual model capacity can be compensated by the model decorrelation.

\subsubsection{Loss Ablation}

\begin{table}
\begin{center}
\caption{Ablation study for different values of $\nu$}
\label{lossablation}
\begin{tabular}{|c|c|c|c|c|c|}
\hline
$\nu$ & 0 & 0.1  & 1 & 5 & 10 \\
\hline
3D err (deg) & 4.51 & 4.46 & \textbf{4.35} & 4.47 & 4.61\\
\hline
\end{tabular}
\end{center}
\end{table}

If we train the models with more epochs (e.g. 10 epochs per cross-validation fold), we obtain an angular error of 4.31/4.71 on MPII/UT multiview database with a Rh-Rh-Fc ensemble. By contrast, we obtain 4.29/4.51 respectively on MPII and UT multiview with Rh-Ou-Fc. Plus, we validated the hyper parameter $\nu$ by testing several values ranging from $0.1$ to $10$. Results are shown in Table \ref{lossablation}. We observe that except for extreme values (e.g. $\nu=10$) the proposed combinatorial loss systematically improves the performances of the algorithm.

\begin{table}
\begin{center}
\caption{Loss ablation with a Rh-Ou-Fc architecture trained for 10 epoch for each cross-validation fold.}
\label{tab:ablloss}
\begin{tabular}{|c|c|c|c|}
\hline
&\multicolumn{2}{c|}{MPII} &  UT multiview \\
\hline
Loss        & 2D err (mm) & 3D err (deg) &  3D err (deg) \\
\hline
\hline
without $\mathcal{L}_{\text{comb}}$& 27.12&4.29&4.51\\
 with $\mathcal{L}_{\text{comb}}$& \textbf{25.43}&\textbf{4.26}&\textbf{4.35}\\
\hline
\end{tabular}
\end{center}
\end{table}

Table \ref{tab:ablloss} shows a comparison of models trained with ($\nu=1$) and without our combinatory loss. As a result, it allows to enhance the predictions on both datasets by forcing each individual combination of models to be relevant for gaze estimation, while at the same time promoting diversity with the random weighting scheme, as it will also be illustrated by introspecting the models.


\begin{figure*}
\begin{minipage}[c]{.49\linewidth}
 \includegraphics[width = \linewidth]{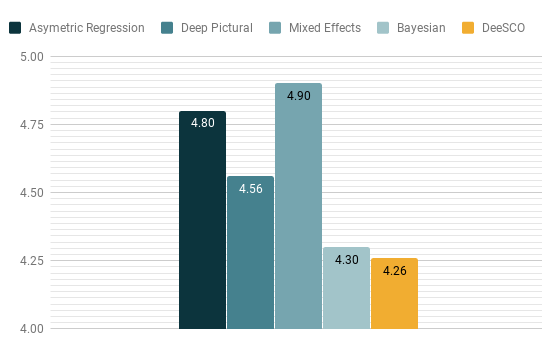}
\caption{Comparisons on MPII in terms of 3D angular error.}
\label{MPII_SOA}
\end{minipage} \hfill
\begin{minipage}[c]{.5\linewidth}
\includegraphics[width = 0.98\linewidth]{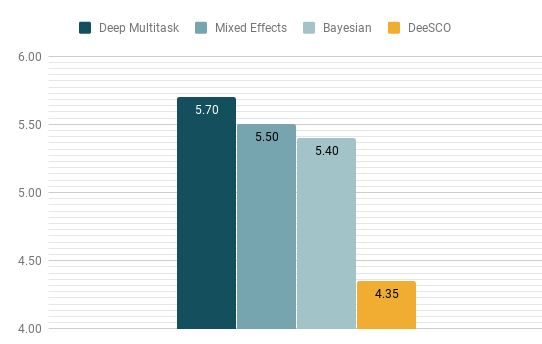}
\caption{Comparisons on UT Multiview in terms of 3D angular error.}
\label{UT_SOA}
\end{minipage}
\end{figure*}

\subsubsection{Comparison with state-of-the-art approaches}\label{compsota}

Figure \ref{MPII_SOA} draws a comparison between our DeeSCo model, which consists in a Rh-Ou-Fc ensemble trained with stochastic combinatory loss, and recent state-of-the-art approaches on MPII dataset. DeeSCo significantly outperforms both asymetric regression \cite{cheng2018appearance} and mixed effects neural networks \cite{xiong2019mixed} and outperforms the deep pictorial gaze estimation proposed in \cite{park2018deep}. It also slightly outperforms the current state-of-the-art approach \cite{wang2019generalizing}, which uses adversarial domain modelization along with a bayesian optimization framework.

Figure \ref{UT_SOA} summarizes results obtained on UT multiview dataset with the same network and learning procedure (Rh-Ou-Fc trained with the proposed combinatory loss), DeeSCo outperforms the work in \cite{wang2019generalizing} by $20\%$, which already outperforms other state-of-the-art approaches \cite{yu2018deep,xiong2019mixed}. Thus, thanks to the use of the heterogeneous ensemble as well as the proposed combinatory loss, DeeSCo sets a new state-of-the-art on both datasets.


\subsection{2D gaze estimation}\label{2dgaze}

We also evaluated our DeeSCo model for 2D gaze estimation on MPII database (Table \ref{tab:2D_models_SOA}). As for 3D gaze estimation we implemented a leave-one-subject-out protocol, returning the average error in millimeters across the 15 folds, as in \cite{sugano2017s}. While state-of-the-art approaches \cite{krafka2016eye,krafka2016eye} have errors in the $46-42 \text{mm}$, DeeSCo achieves \textbf{25.43} accuracy, making a more than $40\%$ error reduction. This shows that our heatmap-based deep heterogeneous ensembles are suited for both 2D/3D gaze estimation. 

\begin{table}
\caption{Comparison with state-of-the-art approaches for 2D gaze. \textbf{Ba:} baseline branch, \textbf{Ou:} Outer product branch, \textbf{Rh:} reshaped heatmap branch, \textbf{Fc:} Fully convolutional branch.}
\label{tab:2D_models_SOA}
\begin{center}
\begin{tabular}{|c|c|c|c|c|}
\hline
 Branch 1         & Branch 2        & Branch 3        & err (mm)          & \#params  \\
\hline\hline
\multicolumn{3}{|l|}{iTracker  \cite{krafka2016eye}} & 45.7 &-\\
\multicolumn{3}{|l|}{iTracker (AlexNet)  \cite{krafka2016eye}} & 46.1 & 62.3M \\
\multicolumn{3}{|l|}{Spatial weights CNN  \cite{sugano2017s}} & 42.0 &-\\
\hline\hline
Ba                & -               & -               & 42.75             & 6.1M  \\
Ou                & Ou              & -               & 38.01             & 13M   \\
Rh                & Ru              & -               & 34.90             & 29M   \\
Fc                & Fc              & Fc              & 30.94             & 1.5M  \\
Ba                & Fc              & -               & 29.46             & 7.6M  \\ 
Ou                & Ou              & Fc              & 30.86             & 14.5M \\
Ou                & Ou              & Ou              & 28.77             & 19.5M \\
Rh                & Rh              & Fc              & 29.45             & 30.5M \\
Rh                & Rh              & Rh              & 28.93             & 43.5M \\
Rh                & Ou              & Fc              & 27.12           & 22.3M \\
\hline
\multicolumn{3}{|l|}{DeeSCo (Rh-Ou-Fc with $\mathcal{L}_{\text{comb}}$)}             & \textbf{25.43}    & 22.3M \\
\hline
\end{tabular}
\end{center}
\end{table}

\subsection{Model introspection}\label{modelintro}

\begin{figure*}
\centering
\includegraphics[width = \linewidth]{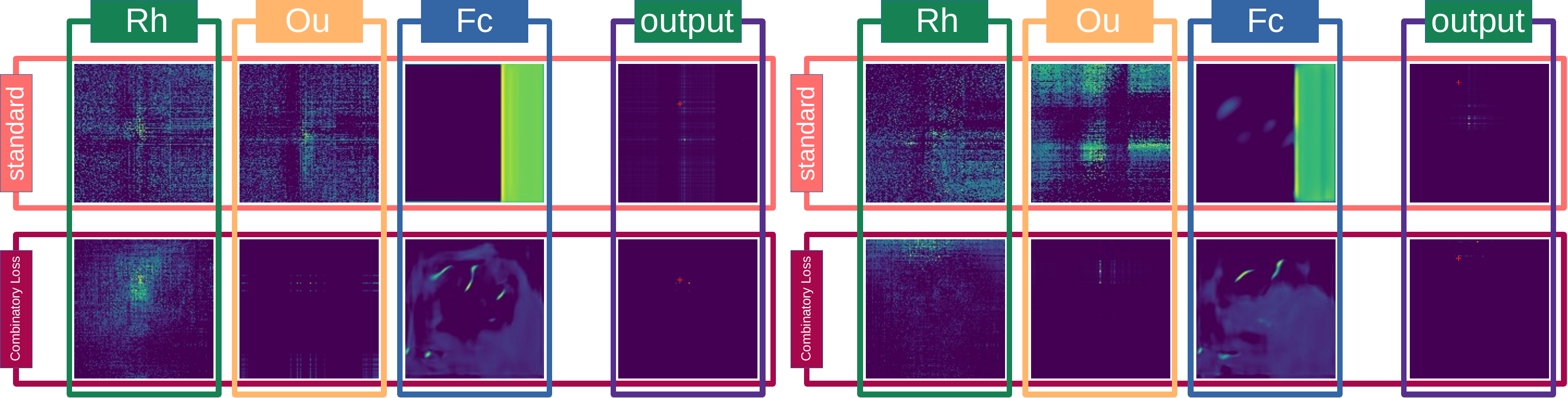}
\caption{Illustration of heatmaps (raw heatmaps, before spatial softmax) generated with a Rh-Ou-Fc heterogeneous ensemble trained with and without the combinatory loss, and combination thereof (after spatial softmax). The red mark indicates the ground truth localization in yaw-pitch space (zoom in to see the difference).}
\label{heatmaps_comparaison_combinatory}
\end{figure*}

Figure \ref{heatmaps_comparaison_combinatory} qualitatively illustrates heatmaps generated by each model of a DeeSCo model (with a Rh-Ou-Fc deep ensemble) with and without our combinatory loss. On the one hand, notice how certain individual weak model do not output qualitatively interesting predictions: for instance, the Fc model, and to a lesser extent the Ou model, only provides insight under the form of a bias in the final prediction. As a result, the output merged heatmap appears as less clear-cut. On the other hand, with our combinatory loss, more emphasis is put so that each separate weak model (and combination thereof) outputs a qualitatively more accurate prediction. Thus, the predictions of the Fc and Ou models are more standalone, yet also satisfyingly combine each other, as they appear closer to the ground truth value, as indicated by the red mark. As a result, the output prediction heatmap exhibits less spatial variance, enhancing the overall accuracy.

\section{Discussion and conclusion}\label{concl}

In this paper, we investigated the use of deep ensemble models for 2D/3D heatmap-based appearance-based gaze estimation. First we show that heterogeneity in the architectural design of the weak predictors that compose the deep ensemble allows to further decorrelate the predictions of these weak predictors, enhancing the predictions of the ensemble as a whole. As such, we proposed three distinct architectural branches, namely a reshaped heatmap branch, a fully-convolutional branch, and an outer product branch. The predictions outputted by each branch are weighted to provide the final prediction of the deep ensemble. During training, the prediction of each of these branches, as well as the weights of the mixture, are learned jointly in an end-to-end manner.
Furthermore, we propose a novel stochastic combinatory loss, which ensures that each combination of weak predictors provides a relevant gaze estimation on its own. This idea, along with random sampling of the combination weights at train time, contributes to provide more robust individual weak predictors as well as to bring extra decorrelation between them. This, in turn, significantly enhances the deep ensemble predictive capacity. Through extensive empirical validation, we show that the proposed deep heterogeneous ensemble trained with stochastic combinatory loss (DeeSCo) significantly outperforms state-of-the-art approaches for both 2D and 3D gaze estimation on several datasets.

The ideas introduced in this paper are however not limited to gaze estimation and can find several applications in adjacent domains. As such, architecturally heterogeneous ensembles could be applied to e.g. ImageNet classification, semantic segmentation, object detection or any other domain where data is abundant and using relatively small ensembles of very deep networks is a solution to enhance the accuracy, beyond going deeper. In such a case, the proposed stochastic combinatory loss could also be applied in a rather straightforward manner. As for gaze estimation, another future direction would be to further decorrelate the weak predictors, e.g. by applying different regularizations (e.g. dropout) at train time. 

{\small
\bibliographystyle{IEEEtran}
\bibliography{egbib}
}

\end{document}